\documentclass{article}

\usepackage{babel}
\usepackage{natbib}
\usepackage[utf8]{inputenc} 
\usepackage[T1]{fontenc}    
\usepackage{hyperref}       
\usepackage{url}            
\usepackage{booktabs}       
\usepackage{amsfonts}       
\usepackage{nicefrac}       
\usepackage{microtype}      
\usepackage{xcolor}         
\usepackage{graphicx}
\usepackage{authblk}
\usepackage{fullpage}

\title{GPT Editors, Not Authors: The Stylistic Footprint of LLMs in Academic Preprints}

\author{
  Soren DeHaan\\
  \vspace{-0.4cm}
  Department of Computer Science\\
  Indiana University, Bloomington \\
  \texttt{sodeha@iu.edu}
  \and
   Yuanze Liu\\   \vspace{-0.4cm} Cognitive Science Program\\ Indiana University, Bloomington\\
  \texttt{yl220@iu.edu}
  \and Johan Bollen \\    \vspace{-0.4cm} Department of Informatics\\
  Indiana University, Bloomington \\ \texttt{jbollen@iu.edu}
  \and Sa\'ul A. Blanco \\   \vspace{-0.4cm} Department of Computer Science\\
  Indiana University, Bloomington \\ \texttt{sblancor@iu.edu}
}

\date{May 22, 2025}

\begin{document}

\maketitle

\begin{abstract}
The proliferation of Large Language Models (LLMs) in late 2022 has impacted academic writing, threatening credibility, and causing institutional uncertainty. We seek to determine the degree to which LLMs are used to generate critical text as opposed to being used for editing, such as checking for grammar errors or inappropriate phrasing. In our study, we analyze arXiv papers for stylistic segmentation, which we measure by varying a PELT threshold against a Bayesian classifier trained on GPT-regenerated text. We find that LLM-attributed language is not predictive of stylistic segmentation, suggesting that when authors use LLMs, they do so uniformly, reducing the risk of hallucinations being introduced into academic preprints.
\end{abstract}

\section{Introduction}

``When ChatGPT was released to the public on November 30, 2022, it quickly changed the landscape of digital communication. In academic circles, this shift has provoked serious concerns about authenticity and trust. The potential use of large language models in producing scholarly content raises red flags—not only about authorship, but about accuracy itself. Hallucinations and false references generated by LLMs represent a genuine risk to the integrity of the academic record.'' [Text generated by ChatGPT].

The prevalence of LLM use in scholarly writing, as in the preceding paragraph, is a hot topic of debate.
Previous works have given an estimate of greater than 1\% of all articles using LLM assistance, with a steady increase after the release of ChatGPT \citep{gray_chatgpt_2024}; however, the percentage estimate of LLM use in research articles varies based on the LLM detection tools used, the databases analyzed, and the areas of research, with some sub-population estimates exceeding 20\% \citep{liang_mapping_2024, picazo-sanchez_analysing_2024, cheng_have_2024}. The use of LLM in the peer review process is also a concern. Existing work has suggested that between 6.5\% and 16.9\% of peer reviews for AI conferences could contain substantial LLM modification (usage beyond spell-checking and minor editing) \citep{liang_monitoring_2024}; AI-assisted peer review is also more likely to recommend acceptance to conferences, undermining trust in the equitability of the review process \citep{latona_ai_2024}.

One danger that LLMs pose to the scientific community is the accuracy of the generated content, particularly due to LLM hallucinations. In Natural Language Generation, hallucination describes the behavior of a language model in generating text that is nonsensical or unfaithful to the provided input \citep{ji_survey_2023}. The hallucinatory behaviors of large language models are of significant concern. Such behaviors include providing inaccurate summarizations of text, information conflicting with context, and non-existent reference articles, all instances of the fact-conflicting hallucinations \citep{tang_tofueval_2024, kim_fables_2024, donker_dangers_2023, zhang_sirens_2023, walters_fabrication_2023, ariyaratne_comparison_2023}. For a research paper, the fabrication of its results due to LLM hallucination is a challenge to detect during the peer review process, and the ramifications are extreme for both the human author(s) and the scientific community \citep{zheng_chatgpt_2023}.

An accurate assessment of the usage of LLM in manuscript preparation, which informs journal policy, is important, as unchecked use of LLM threatens the credibility of scientific publication, while excessive limits on the use of LLM disproportionately affect researchers whose first language is not English. Given how English is used as the common language of science, language poses a significant barrier to the contribution and career development of scholars who are non-native English speakers; more labor is required to access and write articles in English or to deliver presentations at conferences conducted in English \citep{amano_manifold_2023}. LLM detectors are more likely to mislabel non-native English writing as AI generated, unfairly disadvantaging non-native English speakers \citep{liang_gpt_2023}. Yet LLMs can support authors with text readability: Non-native English speakers face more paper rejections explicitly due to writing issues than native English speakers; by harnessing the linguistic capabilities of LLMs in responsible and effective ways, the language barrier within the scientific community can be lowered to allow for greater collaboration \citep{lenharo_sciences_nodate, lin_techniques_2025}. 

Given the possibility of fact-conflicting hallucinations, it is well accepted that critical stages such as writing manuscripts and peer review should not be tasks for generative AI; however, summarization, grammar and spelling checks are low-risk uses of generative AI and can be helpful in scientific research \citep{bockting_living_2023}.


In this paper, we consider four general patterns of LLM use:

\begin{enumerate}
    \item No LLM use, where all text is created by human authors;
    \item Using LLMs for editing or translation, where all base text is still human-authored and the LLM does not significantly mutate conveyed content;
    \item Partial LLM generation, where most of the text is human-authored, but some sections are LLM-generated; and
    \item Complete LLM generation, where all text is generated by an LLM without significant human input.
\end{enumerate}

Assessing whether LLMs are used is difficult at scale, and determining whether the role of the LLM in the writing process was editing or complete generation is an open problem: work is ongoing to successfully distinguish between the second and fourth cases. Significant research has been done to develop stronger LLM detection, distinguishing between the first case -- no LLM use -- and all remaining cases. Existing methods of LLM usage detection include models that use contrastive or probabilistic techniques such as Binoculars and DetectGPT \citep{hans_spotting_2024, mitchell_detectgpt_2023}. Word frequency changes also reflect potential LLM usage, as certain keywords are disproportionally used by LLMs \citep{kobak_delving_2025}.
However, techniques such as paraphrasing and prompt tuning challenge many of the current LLM detectors, suggesting the general intractability of combating adversarial techniques at scale \citep{krishna_paraphrasing_2023, kumarage_how_2023, lu_large_2024, sadasivan_can_2025}.

The goal of this paper is to distinguish between the second and third cases on a population level. In other words, when a paper involves both human writing and LLM generation, can we identify whether the LLM was disproportionately applied to certain parts of the text?

We first train and use a Bayesian classifier to determine the likelihood that each paper has been modified by an LLM. We also use Pruned Exact Linear Time (PELT), a changepoint detection algorithm, to determine how significantly the frequency of LLM-associated words changes over the course of the document. By comparing these results, we can determine whether the use of LLMs corresponds to an increase in human-LLM stylistic variability. If it does not, then that suggests that LLM use is uniform, as would occur when an author uses LLMs for general editing or translation, or when a manuscript is completely fabricated. Our results suggest that there is no such correlation, which implies that-- after accounting for the possibility of complete fabrication-- almost all LLM use is for editing or translation.

\subsection*{Contributions}

Our contributions are a hybrid of literary synthesis, methodological development, and empirical findings. In our synthesis, we enumerate four central patterns of LLM use and non-use, and identify a greater focus on the disambiguation of certain LLM use patterns. Methodologically, we introduce a means of measuring stylistic segmentation by integrating PELT with Bayesian classifiers. Finally, our results show that LLM use tends to be stylistically uniform: authors which leverage LLMs for a portion of a paper tend to use LLMs in the paper's entirety.

\section{Methodology}

\subsection{Data \& Generated Text}

Our data were sourced from the arXiv \citep{arxiv_org_submitters_2024}. We took one uniform sample of preprints on the arXiv from May 2021 to September 2021 ($n=550$), which predates widespread public access to LLMs. These would be used as the basis for the Bayesian classifier, so taking data prior to the public release of ChatGPT all but eliminates the possibility of an LLM-generated text being used in the human training set.

We also took a uniform sample from January 2023 to May 2025 ($n=8000$), on which the brunt of our analysis would focus. There were two potential concerns with sampling: the first is that papers published in early 2023 may not have adapted to the presence of LLMs so may not contain relevant LLM-generated content \citep{liang_mapping_2024}. The second is that papers published more recently may experience a greater divergence in LLM models used, making linguistic analyses challenging. We chose a uniform sample as there is not yet consensus on the most effective assessment range.

In order to prepare the data, we applied nine criteria in order to find section headers. If we found an Abstract header, an Introduction header, and a Conclusion or Discussion header, and if each of the demarcated sections was at least 500 characters, we included the resulting sections in our datasets. This reduced our 2021 dataset to $n=205$ papers and our main dataset to $n=2408$ papers.

The decision to restrict our search to papers containing each of the three sections (Abstract, Introduction, and Conclusion/Discussion) requires justification. The primary reason was that these sections tend to see higher rates of LLM use, due to their consistent format \citep{liang_mapping_2024}. Furthermore, these sections are common across disciplines and are more readily delineated than the methods or results sections. Lastly, papers which contain all three sections are more likely to have consistent formatting throughout, which decreases the risk of preprocessing errors.

Once we restricted our dataset, we then cleaned each section, removing items such as keyword lists, web links, and journal numbers. For each paper, these three sections were concatenated into one combined text file.

For each of the paper sections in the 2021 dataset, we used the OpenAI API to regenerate the texts using GPT 3.5 Turbo. These regenerated texts, along with their original forms, were set aside. Two-thirds of the pairs were used to train the Bayesian classifier. One-sixth of the pairs were used to validate the classifier's accuracy. The remaining sixth were used to validate the PELT segmentation metric against a ground-truth model.

\subsection{Bayesian Classifier}

We trained a naive Bayesian classifier over the words in the original and GPT-regenerated texts. For each document in both the human and LLM data, we computed the frequency rate of each word to account for length bias, and then took the average frequency rate across all documents in their respective datasets. We then converted the odds into a ratio between the human and LLM datasets and cast to a smoothed logarithm to avoid $\log{0}$ errors, as follows for a given word $W$:

\[LogOdds(W) = \log{\frac{\sum LLMOdds(W)+10^{-4}}{\sum HumanOdds(W)+10^{-4}}},\]

where $LLMOdds$ is the average frequency rate across LLM-generated texts and $HumanOdds$ is the average frequency rate in the base texts. For each document, we then applied $LogOdds$ to each word in the text, resulting in a list of log odds. We also generated a cumulative sum over the course of each document. While this method is naive, it is fast and applicable for generating word-by-word probabilities, which is a necessary element for PELT application. Furthermore, Bayesian approaches to classification have been shown to be predictive in the aggregate \citep{liang_monitoring_2024, zhang_enhancing_2024}.

\subsection{PELT \& Segmentation}

Pruned Exact Linear Time (PELT) is a changepoint detection algorithm that specializes in detecting multiple changepoints within a text. It yields more accurate results than binary segmentation, and runs many times faster than optimal partitioning \citep{killick_optimal_2012}. We ran PELT on the classifier-generated log odds cumulative sum arrays, in order to identify whether there was a characteristic shift in the use of LLM-attributed language throughout a text. To account for the inherent scaling effect of variance, we normalized the PELT penalty constant so that it scaled linearly with the variance of the word log odds of each individual document.

While PELT can generate the list of changepoints, we wanted to know how much certainty to take in the validity of there existing such a changepoint. Accordingly, after accounting for the variance of the documents, we computed a penalty multiplier for each document. This was done iteratively, first doubling the multiplier until no changepoints were found, then performing a binary search to isolate the multiplier threshold over which no changepoints were found. This guaranteed a threshold with a margin of error of $10^{-2}$ for a threshold $M$, while running in $O(\log{M})$ PELT passes. PELT has linear runtime with respect to the length $n$ of the document, giving a $O(n\log{M})$ runtime \citep{killick_optimal_2012}.

In this way, the threshold multiplier acts as a variance-normalized metric of how much noise we would need to introduce into the system to disguise any segmentations within the text. This means that the multiplier captures the degree to which the LLM-associated words within a paper are clumped together, as the more tightly the words are grouped, the more noise would be required to disguise the segmentation.

\subsection{Validation}

\begin{figure}
   \begin{minipage}{0.49\textwidth}
    \centering
    \caption{Histogram of classifier results on GPT-regenerated text.}
    \includegraphics[width=\linewidth]{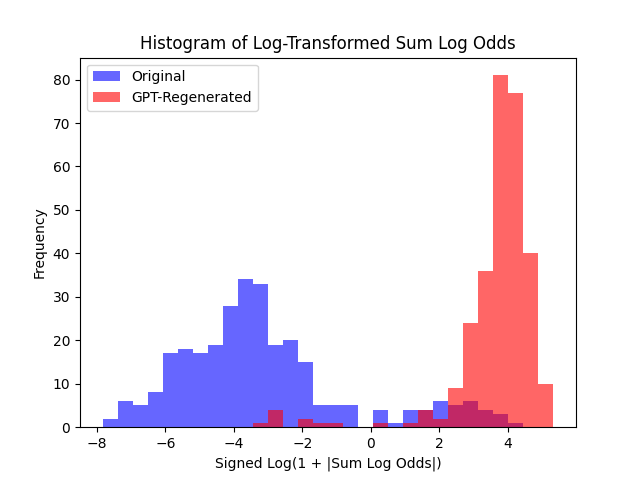}
    \label{fig:log_hist}
   \end{minipage}\hfill
   \begin{minipage}{0.49\textwidth}
    \centering
    \caption{Confusion matrix for the classifier.}
    \includegraphics[width=\linewidth]{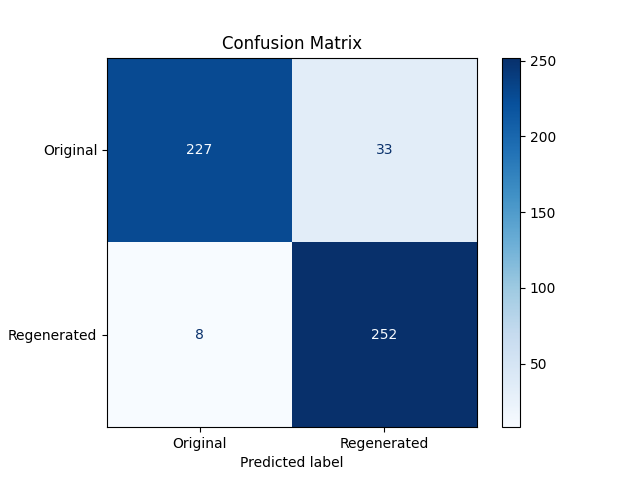}
    \label{fig:log_confusion}
   \end{minipage}\hfill
\end{figure}

In order to validate the classifier, we used the remaining 2021 data (one third of the total train/test set) to test the classifier's accuracy. For each of the individual demarcated sections (increasing the sample to $n=260$), we again regenerated the text using OpenAI's API, then ran the classifier over both the LLM-generated documents and the original human documents. The results are described as both a logarithmic histogram in Figure~\ref{fig:log_hist} and a confusion matrix in Figure~\ref{fig:log_confusion}.

We used that same third of the 2021 data to verify our PELT threshold technique. To do this, we regenerated the text as before, but also created a segmented sample, using OpenAI to add an entirely new paragraph to the existing text. This gave us three datasets: the completely LLM-generated text, the original human text, and the human text with an LLM-generated addition. On average, the LLM addition was $19.4\%$ of the length of the human text.

\begin{figure}
    \centering
    \caption{The distribution difference of PELT thresholds for the segmented dataset.}
    \includegraphics[width=0.8\linewidth]{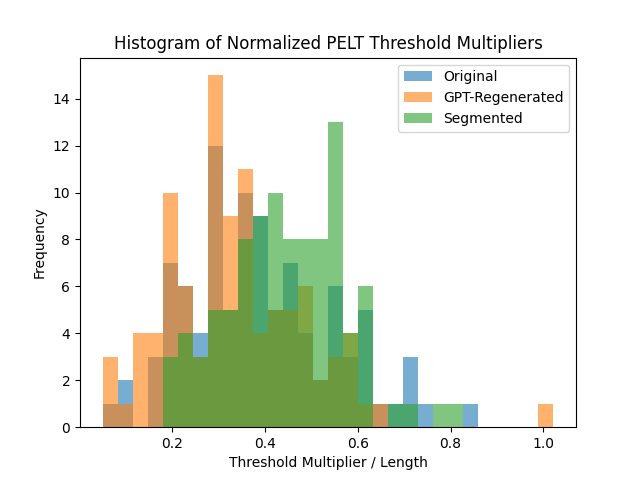}
    \label{fig:threshold_hist}
\end{figure}

For each of these three datasets, we ran the Bayesian classifier, then computed their PELT threshold, normalized for length, and compared the results. Tables \ref{table:PELT_verify_meansd} and \ref{table:PELT_likelihoods} give population-level statistics, and the results are visualized in Figure~\ref{fig:threshold_hist}. The distinctions are clearly differentiable in the aggregate, even with less than $20\%$ LLM use. Critically, the segmented dataset has a higher PELT threshold multiplier than both the original human texts and the LLM-regenerated texts ($p < 0.01$), demonstrating the validity of this technique.

\begin{table}[ht]
\centering
\caption{Summary statistics of classification scores by group.}
\begin{tabular}{lcc}
\hline
\textbf{Group} & \textbf{Mean} & \textbf{Standard Deviation} \\
\hline
Original       & 0.3916 & 0.1595 \\
Regenerated    & 0.3363 & 0.1499 \\
Segmented      & 0.4471 & 0.1269 \\
\hline
\end{tabular}
\label{table:PELT_verify_meansd}
\end{table}

\begin{table}[ht]
\centering
\caption{Pairwise comparison of group likelihoods. The segmented group is strongly differentiated from both of the other groups.}
\begin{tabular}{lcc}
\hline
\textbf{Comparison} & \textbf{t-statistic} & \textbf{p-value} \\
\hline
Original vs. Regenerated &  2.488 & $< 0.05$ \\
Segmented vs. Original   &  2.678 & $< 0.01$ \\
Segmented vs. Regenerated&  5.551 & $< 0.001$ \\
\hline
\end{tabular}
\label{table:PELT_likelihoods}
\end{table}

\section{Results}

\subsection{Qualitative Classifier Analysis}

The classifier identified a large set of words as being particularly indicative of either human authorship or LLM authorship. We observed that significantly more words were considered indicative of LLM authorship than were indicative of human authorship. Below is a sampling of each:

\begin{table}[h!]
\centering
\caption{LLM-associated words, log odds > 5.6}
\begin{tabular}{lr}
\textbf{Word} & \textbf{Value} \\
\hline
confidence & 6.71 \\
efficiency & 6.68 \\
configuration & 6.66 \\
utilization & 6.50 \\
efficacy & 6.36 \\
floating & 6.17 \\
efficiently & 6.16 \\
offering & 6.14 \\
significance & 6.12 \\
endeavor & 6.07 \\
verification & 6.06 \\
beneficial & 6.04 \\
realm & 6.00 \\
significant & 5.99 \\
unified & 5.92 \\
classification & 5.88 \\
synchronization & 5.86 \\
overflow & 5.75 \\
affecting & 5.73 \\
examination & 5.72 \\
delf (delve) & 5.71 \\
specific & 5.69 \\
artificial & 5.66 \\
innovative & 5.64 \\
modification & 5.63 \\
\end{tabular}
\end{table}

\begin{table}[h]
\centering
\caption{Human-associated words, log odds > 5.6. Note the shorter length.}
\begin{tabular}{lr}
\textbf{Word} & \textbf{Value} \\
\hline
because & -6.79 \\
come & -6.01 \\
substitute & -5.81 \\
expect & -5.75 \\
cent (recent) & -5.75 \\
shall & -5.74 \\
devise & -5.71 \\
thank & -5.65 \\
\end{tabular}
\end{table}

Some general observations hold. LLM-associated words tend to be more ambiguous in their meaning and are often longer and more flowery than their human-associated counterparts. We also note that human-associated words include more formal restrictive words, such as ``because'' and ``substitute'', affective words, such as ``thank'' and ``great'', and words that refer to a particular time, such as ``recent'' and ``nowadays''. These results are generally in line with previous studies and align with our expectations, but are provided for completeness \citep{kobak_delving_2025}.

\subsection{Segmentation Correlations}

\begin{figure}[h!]
    \centering
    \caption{The original data. Observe length as a confounding variable.}
    \includegraphics[width=0.8\linewidth]{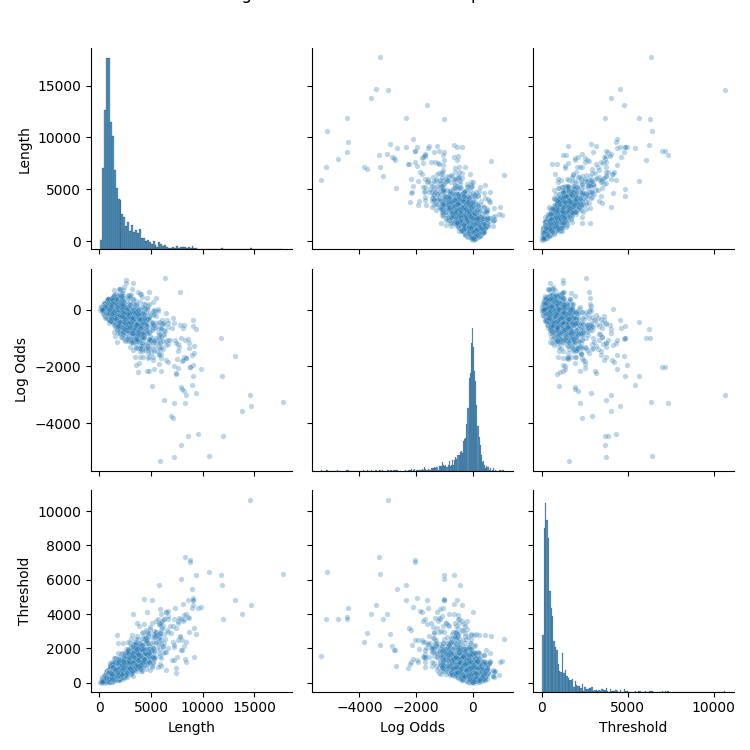}
    \label{fig:original}
\end{figure}

The total log-odds of a given paper -- that is, the confidence with which the classifier predicts the paper is LLM generated -- correlates with the PELT multiplier, but both scale close to linearly with the length of the text (see Figure~\ref{fig:original}). The total log odds has a magnitude that scales with text length as well, but it is apparent that length is a confounding variable for both. For precise correlations, see Table~\ref{table:no_correlation}.

\begin{figure}
    \centering
    \caption{After normalization with Z-score.}
    \includegraphics[width=0.8\linewidth]{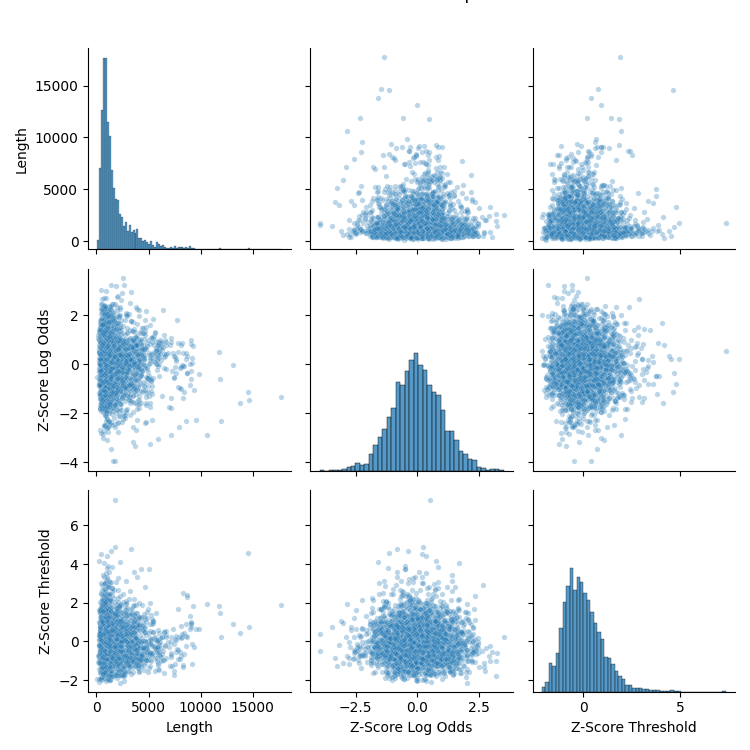}
    \label{fig:zscore}
\end{figure}

In order to account for the confounding length variable, we apply Z-score normalization with $25$ bins, so that each total log-odds and PELT threshold is normalized with respect to a neighborhood of lengths, eliminating the correlation between both variables and the confounding length variable. However, after applying this normalization, we find that the original correlation between total log-odds and PELT multipliers is gone (see Table~\ref{table:no_correlation}). Note that this normalization was only applied for each variable with respect to length. No normalization was applied for total log-odds with respect to the PELT threshold. This implies that the original correlation between PELT threshold and total log-odds was only a result of the confounding length variable and that there is no correlation between the two. A visualization of the normalized data can be found in Figure~\ref{fig:zscore}.

\begin{table}[h]
\centering
\caption{Pearson Correlation Coefficients and p-values. Note how Z-score normalization with length removes all statistical significance.}
\begin{tabular}{lrr}
\toprule
\textbf{Variable Pair} & \textbf{Correlation ($r$)} & \textbf{p-value} \\
\midrule
\multicolumn{3}{c}{\textit{Original Variables}} \\
Length vs. Log Odds & $-0.7204$ & $< 0.001$ \\
Length vs. Threshold & $0.8806$ & $< 0.001$ \\
Log Odds vs. Threshold & $-0.6409$ & $< 0.001$ \\
\midrule
\multicolumn{3}{c}{\textit{Z-Score Normalized Variables on Length}} \\
Length vs. Z-Score Log Odds & $-0.0256$ & $> 0.05$ \\
Length vs. Z-Score Threshold & $0.0326$ & $> 0.05$ \\
Z-Score Log Odds vs. Z-Score Threshold & $-0.0233$ & $> 0.2$ \\
\bottomrule
\end{tabular}
\label{table:no_correlation}
\end{table}

\subsection{Section Correlations}

As a supporting metric for determining whether LLM use is uniform, we compute the correlation in LLM log odds between each of the individual paper sections. As expected from the uniformity results in our segmentation analysis, we have high correlation between the log odds for each of the sections (see Table~\ref{table:correlation_matrix}).

\begin{table}[ht]
\centering
\caption{Pearson correlation coefficients between normalized log-odds scores for different sections. *** $p < 0.001$.}
\begin{tabular}{lcccc}
\hline
\textbf{Section} & \textbf{Abstract} & \textbf{Introduction} & \textbf{Conclusion} & \textbf{Combined} \\
\hline
Abstract     & 1.000 & 0.595*** & 0.556*** & 0.739*** \\
Introduction & 0.595*** & 1.000 & 0.624*** & 0.906*** \\
Conclusion   & 0.556*** & 0.624*** & 1.000 & 0.809*** \\
Combined     & 0.739*** & 0.906*** & 0.809*** & 1.000 \\
\hline
\end{tabular}
\label{table:correlation_matrix}
\end{table}

\section{Limitations}

Several limiting factors restrict the degree to which these results can be generalized. Most immediately, our data were taken exclusively from the arXiv, and may not be a representative sample of academic preprints as a whole. Further, as a population level analysis, the lack of signal does not prove that there are no instances of partial LLM generation, merely that they represent a statistically insignificant sample. Caution is still warranted when addressing LLM generated content. Additionally, as this analysis assesses the field quantitatively, the picture won't be complete without a more thorough qualitative study to explain when and how researchers use LLMs.

This study is also limited by the scope of LLMs. While the word distributions of GPT 3.5 Turbo are likely representative of multiple LLMs, there are a very large number of LLMs available to researchers at this time, and not all of them will have the same linguistic biases. As the number and variety of LLMs continues to increase, this type of analysis will become increasingly difficult to perform. In the event that LLMs are banned from use in some journals, enforced by LLM detection programs, we may see a race of adversarial techniques to overcome detection, further complicating this area of study.

Additional limitations include the possibility that LLM use patterns have changed over time or across disciplines. As our sample was taken uniformly across several years, without respect to discipline, we are only addressing population level statistics at the highest level.

\section{Discussion}

Over the course of this paper, we have determined that when LLMs are used in academic preprints, they are uniformly applied. Our method does not distinguish between a text which is generated entirely by an LLM and one which is edited by one, but our results suggest that these collectively are more common use cases. Of the four LLM use cases we have described, the third use case -- using an LLM to generate part of a text -- is highly uncommon.

This should be reassuring to a degree, as in the absence of texts which are completely LLM generated, we have evidence that LLM editing far outweighs (partial) LLM generation. As partial LLM generation has a greater risk of introducing hallucinations than LLM editing, it is warranted to prioritize distinguishing complete LLM generation from LLM editing and translation.

Taken broadly, these results suggest that the increase in LLM use in academic preprints does not imply a proportional increase in LLM generated content. Rather, the increase in the use of LLMs appears to be related to the use of LLMs as editing and translation tools, which may instead improve the clarity and accessibility of research. We must remain vigilant against bad actors and the potential abuse of LLM tools, but in our global research community, these results suggest that researchers have used LLMs responsibly.

To ensure that research remains clear and accessible, it seems advisable for journals and conferences to permit the use of LLMs, provided that their use is properly disclosed \citep{van_dis_chatgpt_2023}. It is critical that submissions labeled as having used LLMs are not seen as necessarily of lower quality, since such an opinion might discourage honest disclosures and potential collaborations with non-native speakers. It is essential to understand the potential risks and benefits associated with LLM use, but also to know that using LLMs does not inherently reduce the quality of the work. LLMs are tools, and it is commitment to academic honesty and collaboration that will ensure that LLM use is beneficial and ethical.

\section{Acknowledgments}

 S. DeHaan and S. Blanco were supported in part by the US Department of Defense [Contract No. W52P1J2093009], through the NSWC Crane SCALE program.

\section*{Code:}




 The codebase used for the project can be found at \hyperlink{https://github.com/sodeha-mirror/gpt-watch}{https://github.com/sodeha-mirror/gpt-watch}.



\bibliographystyle{plainnat}

\end{document}